%% file: 0_main.tex
\documentclass{article}

    \PassOptionsToPackage{numbers, compress}{natbib}

    \usepackage[preprint]{neurips_2025}

\usepackage[utf8]{inputenc} 
\usepackage[T1]{fontenc}    
\usepackage{hyperref}       
\usepackage{url}            
\usepackage{booktabs}       
\usepackage{amsfonts}       
\usepackage{nicefrac}       
\usepackage{microtype}      
\usepackage{xcolor}         
\usepackage{graphicx}
\usepackage{multirow}
\usepackage{threeparttable}
\usepackage{siunitx}
\title{Glucose-ML: A collection of longitudinal diabetes datasets for development of robust AI solutions}

\author{%
  Temiloluwa Prioleau\thanks{Use footnote for providing further information
    about author (webpage, alternative address)---\emph{not} for acknowledging
    funding agencies.} \\
  Department of Computer Science\\
  Emory University\\
  Atlanta, GA 30322 \\
  \texttt{tpriole@emory.edu} \\
  \And
  Baiying Lu \thanks{Equal contribution.}\\
  Department of Computer Science \\
  Dartmouth College \\
  Hanover, NH 03755 \\
  \texttt{baiying.lu.gr@dartmouth.edu} \\
  \AND
  Yanjun Cui \footnotemark[2] \\
  Department of Computer Science \\
  Dartmouth College \\
  \texttt{yanjun.cui.gr@dartmouth.edu} \\
}

\begin{document}

\maketitle

\begin{abstract}
Artificial intelligence (AI) algorithms are a critical part of state-of-the-art digital health technology for diabetes management. Yet, access to large high-quality datasets is creating barriers that impede development of robust AI solutions. To accelerate development of transparent, reproducible, and robust AI solutions, we present Glucose-ML, a collection of 10 publicly available diabetes datasets, released within the last 7 years (i.e., 2018 - 2025). The Glucose-ML collection comprises over 300,000 days of continuous glucose monitor (CGM) data with a total of 38 million glucose samples collected from 2500+ people across 4 countries. Participants include persons living with type 1 diabetes, type 2 diabetes, prediabetes, and no diabetes. To support researchers and innovators with using this rich collection of diabetes datasets, we present a comparative analysis to guide algorithm developers with data selection. Additionally, we conduct a case study for the task of blood glucose prediction - one of the most common AI tasks within the field. Through this case study, we provide a benchmark for short-term blood glucose prediction across all 10 publicly available diabetes datasets within the Glucose-ML collection. We show that the same algorithm can have significantly different prediction results when developed/evaluated with different datasets. Findings from this study are then used to inform recommendations for developing robust AI solutions within the diabetes or broader health domain. We provide direct links to each longitudinal diabetes dataset in the Glucose-ML collection and openly provide our code: \url{https://github.com/Augmented-Health-Lab/Glucose-ML_arxiV2025}.



\end{abstract}

\input{1_introduction}

\input{2_related_work}
\input{3_dataset_overview}

\input{4_comparative_analysis}
\input{5_glucose_prediction}
\input{6_discussion}

\bibliographystyle{acm}
\bibliography{8_references}









\input{9_appendix}


\newpage

\section*{NeurIPS Paper Checklist}

\begin{enumerate}

\item {\bf Claims}
    \item[] Question: Do the main claims made in the abstract and introduction accurately reflect the paper's contributions and scope?
    \item[] Answer: \answerYes{} 
    \item[] Justification: See section 1, 3, 4, and 5.5.

\item {\bf Limitations}
    \item[] Question: Does the paper discuss the limitations of the work performed by the authors?
    \item[] Answer: \answerYes{} 
    \item[] Justification: See section 6.3.

\item {\bf Theory assumptions and proofs}
    \item[] Question: For each theoretical result, does the paper provide the full set of assumptions and a complete (and correct) proof?
    \item[] Answer: \answerNA{} 
    \item[] Justification: This paper does not include any theoretical results that require assumptions or proofs.

    \item {\bf Experimental result reproducibility}
    \item[] Question: Does the paper fully disclose all the information needed to reproduce the main experimental results of the paper to the extent that it affects the main claims and/or conclusions of the paper (regardless of whether the code and data are provided or not)?
    \item[] Answer: \answerYes{} 
    \item[] Justification: See section 4 and 5.

\item {\bf Open access to data and code}
    \item[] Question: Does the paper provide open access to the data and code, with sufficient instructions to faithfully reproduce the main experimental results, as described in supplemental material?
    \item[] Answer: \answerYes{} 
    \item[] Justification: A link to all code created for this paper is provided in the abstract. Additionally, all datasets used in this study are publicly available and a direct link to each dataset in provided in section 3 and Table 1.

\item {\bf Experimental setting/details}
    \item[] Question: Does the paper specify all the training and test details (e.g., data splits, hyperparameters, how they were chosen, type of optimizer, etc.) necessary to understand the results?
    \item[] Answer: \answerYes{} 
    \item[] Justification: See section 5.3.

\item {\bf Experiment statistical significance}
    \item[] Question: Does the paper report error bars suitably and correctly defined or other appropriate information about the statistical significance of the experiments?
    \item[] Answer: \answerYes{} 
    \item[] Justification: See section 5.4 and Fig. 2.

\item {\bf Experiments compute resources}
    \item[] Question: For each experiment, does the paper provide sufficient information on the computer resources (type of compute workers, memory, time of execution) needed to reproduce the experiments?
    \item[] Answer: \answerYes{} 
    \item[] Justification: See section 5.3.
    
\item {\bf Code of ethics}
    \item[] Question: Does the research conducted in the paper conform, in every respect, with the NeurIPS Code of Ethics \url{https://neurips.cc/public/EthicsGuidelines}?
    \item[] Answer: \answerYes{} 
    \item[] Justification: See section 6.2 and 6.3.

\item {\bf Broader impacts}
    \item[] Question: Does the paper discuss both potential positive societal impacts and negative societal impacts of the work performed?
    \item[] Answer: \answerYes{} 
    \item[] Justification: See section 6.1, 6.2. and 6.3.
    
\item {\bf Safeguards}
    \item[] Question: Does the paper describe safeguards that have been put in place for responsible release of data or models that have a high risk for misuse (e.g., pretrained language models, image generators, or scraped datasets)?
    \item[] Answer: \answerYes{} 
    \item[] Justification: See section 3.1, 3.2 and Intended Data Use section in 6.2.

\item {\bf Licenses for existing assets}
    \item[] Question: Are the creators or original owners of assets (e.g., code, data, models), used in the paper, properly credited and are the license and terms of use explicitly mentioned and properly respected?
    \item[] Answer: \answerYes{} 
    \item[] Justification: See section 3.

\item {\bf New assets}
    \item[] Question: Are new assets introduced in the paper well documented and is the documentation provided alongside the assets?
    \item[] Answer: \answerYes{} 
    \item[] Justification: See section 3, 4, 5 and 6.

\item {\bf Crowdsourcing and research with human subjects}
    \item[] Question: For crowdsourcing experiments and research with human subjects, does the paper include the full text of instructions given to participants and screenshots, if applicable, as well as details about compensation (if any)? 
    \item[] Answer: \answerNA{} 
    \item[] Justification: This paper does not involve crowdsourcing or research with human subjects. Human subject means a living individual about whom an investigator conducting research:(i) Obtains information through interaction with the individual; or (ii) Obtains, uses, studies, analyzes, or generates identifiable private information or identifiable biospecimens.

\item {\bf Institutional review board (IRB) approvals or equivalent for research with human subjects}
    \item[] Question: Does the paper describe potential risks incurred by study participants, whether such risks were disclosed to the subjects, and whether Institutional Review Board (IRB) approvals (or an equivalent approval/review based on the requirements of your country or institution) were obtained?
    \item[] Answer: \answerNA{} 
    \item[] Justification: This paper does not involve research with human subjects. This project involves the analysis of publicly available deidentifed data and there is a data user agreement in place for controlled-access datasets.

\item {\bf Declaration of LLM usage}
    \item[] Question: Does the paper describe the usage of LLMs if it is an important, original, or non-standard component of the core methods in this research? Note that if the LLM is used only for writing, editing, or formatting purposes and does not impact the core methodology, scientific rigorousness, or originality of the research, declaration is not required.
    \item[] Answer: \answerNA{} 
    \item[] Justification: This work does not involve LLMs as an important, original, or non-standard component.

\end{enumerate}

\end{document}

%% file: 1_introduction.tex

\section{Introduction}

An artificial intelligence (AI) algorithm is only as good as the data used for development and evaluation. As a result, access to high-quality datasets is a precursor for building robust AI algorithms. This fundamental principle motivates the growing effort to increase access to high-quality datasets and in so doing increase the transparency, reproducibility, generalizability, fairness, and robustness of AI algorithms, especially in human-centered domains like healthcare \cite{rajpurkar2022ai,pineau2021improving,nazer2023bias,chen2023algorithmic,mcdermott2021reproducibility}. Yet, access to large high-quality datasets is a challenge in many health domains where a patients' privacy and security is critical. This challenge motivates the target goal of making large volumes of health-relevant data accessible for research while minimizing potential risks and threats \cite{kostkova2016owns,waring2020automated,pastorino2019benefits}. 

Diabetes is an exemplar health domain in which AI algorithms are a critical part of state-of-the-art technology (e.g. for screening, decision-support, and management \cite{guan2023artificial,contreras2018artificial,ellahham2020artificial,gautier2021artificial,sheng2024artificial,nomura2021artificial}). The prevalence of digital health technology, such as wearable continuous glucose monitors (CGMs), insulin pumps, and activity trackers, is creating large volumes of longitudinal diabetes-relevant data that is critical to develop and evaluate transformational AI-driven solutions. However, progress within the field is hampered by the highly regulated nature of the healthcare industry, low interoperability across technology, and a fragmented research community. Consequently, data-centric researchers and innovators have limited or no access to large high-quality datasets for developing robust AI solutions. 

To bridge this gap, we present Glucose-ML - a collection of longitudinal diabetes-relevant datasets to accelerate development and evaluation of transparent, reproducible, and robust AI solutions that can revolutionize the field. This study focuses on recent diabetes datasets that are publicly available through open or controlled access, and released within the last 7 years (2018 - 2025). All datasets in the Glucose-ML collection include real-world glucose data collected using clinically-validated CGMs across populations of people living with type 1 diabetes (T1D), type 2 diabetes (T2D), pre-diabetes (PreD), and people without diabetes. The key contributions of this work are:
\begin{itemize}
    \item We curate and present Glucose-ML - a collection of 10 publicly-available, diabetes-relevant datasets comprising longitudinal glucose data from over 2500 people across 4 countries, to support robust development and evaluation of AI-driven algorithms, models, and tools. Glucose-ML includes over 300,000 days of CGM data and a total of 38 million blood glucose samples.
    \item We present a comparative analysis of individual datasets within the Glucose-ML collection to highlight strengths and weaknesses of each dataset, and to guide algorithm developers with the dataset selection process when seeking to develop AI solutions.
    \item We present a case study for the common machine learning (ML) task of blood glucose prediction. Through this analysis, we characterize and compare the performance of two na\"{\i}ve prediction algorithms across 10 publicly available diabetes datasets. Additionally, we show that the same model can have significant performance differences for predicting glucose when developed/evaluated on different datasets. 
\end{itemize}

Findings from this study serve as a basis for recommendations to practitioners developing AI solutions for diabetes and within the broader health domain, especially as it relates to data selection, model design, and model evaluation. 


%% file: 2_related_work.tex
\section{Background \& Related Work}
 Wearable CGMs are minimally-invasive sensors for continuous monitoring of glucose trends \cite{bartolome2021glucomine}. Based on today's technology, CGMs provide real-time glucose readings based on sensor measurements from the interstitial fluid \cite{klonoff2017continuous}. Most CGMs record one glucose sample every 5 - 15 minutes depending on the manufacturer, and have a lifespan of 10 - 14 days \cite{klonoff2017continuous}. CGMs are considered to be the standard of care for people living with T1D, and increasing evidence shows the benefit and effectiveness of CGM use in populations of people living with T2D \cite{ajjan2024continuous,lin2021continuous,slattery2017clinical}. Additionally, in recent years, CGM use is becoming more popular in populations of people with prediabetes, people without diabetes, and in sports \cite{bowler2022use,holzer2022continuous,keshet2023cgmap,klonoff2023use}. Other diabetes-relevant wearable devices that monitor and record longitudinal data include insulin pumps used for administering insulin (the hormone needed to metabolize glucose), and activity trackers used for monitoring relevant behavioral and physiological factors, such as physical activity, sedentary behavior, heart rate, body temperature, and sleep metrics \cite{franssen2020can,rodriguez2021mobile}. User-generated data from these digital health technologies is foundational for understanding diabetes management in outpatient settings and developing novel AI solutions to support care \cite{tauschmann2018technology,nwokolo2023artificial,daly2021technology,guan2023artificial,keshet2023cgmap,lu2024mealtime,bartolome2022computational,riddell2023examining}.
 

 While there are existing efforts to increase access to and availability of high quality data in the diabetes domain, the fragmented nature of this interdisciplinary field has created silos between various stakeholders, including data creators and algorithm developers. For example, the Jaeb Center for Health Research (JCHR) - a nonprofit research center that conducts world class clinical trials with a core focus on T1D - provides public access to rich datasets from their studies \cite{jaeb-JCHR}. However, the majority of these datasets are not leveraged within the AI and ML community for algorithmic innovation as evident from several literature reviews \cite{felizardo2021data,woldaregay2019data,yabo2022review}. The most widely used publicly accessible datasets used for diabetes-relevant AI and ML tasks include: the Pima Indian Diabetes Dataset \cite{yabo2022review,chang2023pima,naz2020deep}, UVA/Padova simulated dataset for T1D \cite{man2014uva,visentin2018uva}, and OhioT1DM \cite{marling2020ohiot1dm}. 
 
 The Pima Indian Diabetes Dataset (PIDD) was originally collected by the National
Institutes of Diabetes and Digestive and Kidney Diseases (NIDDK) and hosted on the UCI Machine Learning Database. This relatively small and static PIDD dataset that includes 8 distinct attributes from 768 Pima-Indian women (e.g. age, plasma glucose from a 2-hour oral glucose test, and diastolic blood pressure) has been used for ML research since year 2007 (18 years ago) and is still being used in year 2025 \cite{polat2007expert,reza2024improving,shams2025novel}. Conversely, the UVA/Pavoda Type 1 Diabetes simulator has been released in three versions (S2008 \cite{kovatchev2009silico}, S2013 \cite{man2014uva}, and S2017 \cite{visentin2018uva}) each of which includes virtual T1D subjects (adults, adolescents, and children). The UVA/Pavoda simulated T1D datasets have been used extensively for AI and ML research toward automated insulin delivery systems for closed-loop blood glucose control (also known as an artificial pancreas) \cite{fox2020deep,lee2020toward}. While simulated datasets are critical for developing control algorithms, best practice guidelines show the need for AI/ML algorithms to also be evaluated on real-world data because algorithms evaluated on simulated data can perform poorly in real-world settings \cite{jacobs2023artificial}. 

Amongst the publicly-available, real-world, and diabetes-relevant datasets released in the last 10 years, the OhioT1DM dataset is the most widely in the AI and ML community \cite{marling2020ohiot1dm,felizardo2021data,jacobs2023artificial}. The OhioT1DM dataset was originally published in 2018 and then updated in 2020. In total, this dataset comprises 8-weeks of time-matched CGM and insulin pump data from 12 people with T1D, as well as physiological data from a fitness band/activity tracker. The OhioT1DM dataset has been used extensively for developing and evaluating AI algorithms for blood glucose prediction \cite{zhu2018deep,zhu2022personalized,martinsson2020blood,yang2023short,gu2020neural}. While the OhioT1DM dataset has served as a good benchmark for developing and evaluating AI algorithms, it is limited by the small sample size and limited heterogeneity (e.g., all participants had T1D and used an insulin pump for diabetes management). Given this, it is expected that AI algorithms developed on such a small dataset could performed differently on a larger and more heterogeneous cohort with greater glycemic variability. To bridge the data challenge that stifles development and evaluation of robust AI solutions for diabetes and other time series modeling tasks \cite{wang2022systematic}, we present a collection of publicly-available longitudinal datasets from over 2500 people across multiple countries, including persons with T1D, T2D, pre-diabetes, and no diabetes. 

%% file: 3_dataset_overview.tex
\section{GlucoseML Overview}
Our collection of diabetes datasets - GlucoseML - includes 10 publicly available datasets with longitudinal data from wearable devices, including CGMs, from populations with diabetes (i.e., T1D and T2D) and populations without diabetes (i.e., prediabetes or no diabetes). This collection prioritizes recent datasets, released between year 2018 and 2025, from observational studies and/or retrospective data collections that are not influenced by a clinical intervention. The majority of datasets included in this collection are in alignment with the well-known FAIR data principles, such that each dataset is findable, accessible, interoperable, and reusable \cite{wilkinson2016fair}. Additionally, all datasets include associated documentation that describe the dataset (e.g. through a publication or publicly accessible protocol). 

Table \ref{tab:dataset_overview_update} provides an overview of the Glucose-ML collection which includes longitudinal glucose data from OhioT1DM \cite{marling2020ohiot1dm,OhioT1DM-dataset}, T1DEXI \cite{riddell2023examining}, BIG IDEAs \cite{bent2021engineering,BIGIDEAs-dataset}, DiaTrend \cite{prioleau2023diatrend,DiaTrend-dataset}, ShanghaiT1DM \cite{zhao2023chinese,ShanghaiDM-dataset}, ShanghaiT2DM \cite{zhao2023chinese,ShanghaiDM-dataset}, T1DiabetesGranada \cite{rodriguez2023t1diabetesgranada,T1DiabetesGranada-dataset}, AI-READI \cite{ai2024ai,AIREADI-dataset}, UCHTT1DM \cite{langarica2024deep,UCHTT1DM-dataset}, and CGMacros \cite{CGMacros-dataset}. From Table \ref{tab:dataset_overview_update}, we can observe the collection includes diabetes datasets collected in the U.S. (60\%), China (20\%), Spain (10\%), and Chile (10\%). There are a total of 2559 participants across the collection of datasets, between 12 - 1067 participants in each dataset, and a duration of 2 - 1907 days of glucose data per participant. Across the collection of datasets, there is a total of 313,043 days with at least 1 glucose reading from a wearable CGM and a total of 38,008,810 (i.e., over 38 million) glucose readings. More details about the population breakdown, age, hemoglobin A1C, and summary statistics on glycemic control for each individual dataset is presented in Appendix Table \ref{tab:detailed-overview-diabetes-datasets} and \ref{tab:BGcontrol-metrics}.

\begin{table}[ht]
  \caption{Overview of the Glucose-ML collection comprising 10 public diabetes datasets available for development of robust AI solutions.}
  \label{tab:dataset_overview_update}
  \centering
  \resizebox{\textwidth}{!}{%
  \begin{tabular}{lllllllll}
    \toprule
    \multirow{2}{*}{\shortstack{Publicly available \\Diabetes Datasets}} & \multirow{2}{*}{Accessibility} & \multirow{2}{*}{Year} & \multirow{2}{*}{Country} & \multirow{2}{*}{\shortstack{Person \\Count}} & \multicolumn{2}{c}{Days / Person} & \multirow{2}{*}{\shortstack{Total Glucose\\ Samples}} \\
    \cmidrule(lr){6-7}a
     & & & & &  Mean & Range & \\
    \midrule
    OhioT1DM \cite{OhioT1DM-dataset,marling2020ohiot1dm} & Controlled & 2020$^{1}$ & US & 12  & 54 & 48--57 & 166,533 \\
    T1DEXI \cite{T1DEXI-dataset,riddell2023examining} & Controlled & 2022 & US & 497  & 27 & 3--28 & 3,785,253 \\
    BIG IDEAs \cite{BIGIDEAs-dataset,bent2021engineering} & Open & 2023 & US & 16  & 9.38 & 8--11 & 36,898 \\
    DiaTrend \cite{DiaTrend-dataset,prioleau2023diatrend} & Controlled & 2023 & US & 54  & 512 & 31--1907 & 7,680,740 \\
    ShanghaiT1DM \cite{ShanghaiDM-dataset,zhao2023chinese} & Open & 2023 & China & 12  & 15 & 7--42 & 15,695 \\
    ShanghaiT2DM \cite{ShanghaiDM-dataset,zhao2023chinese} & Open & 2023 & China & 100 & 13 & 4--41 & 112,475 \\
    T1DiabetesGranada \cite{T1DiabetesGranada-dataset,rodriguez2023t1diabetesgranada} & Controlled & 2024 & Spain & 736 & 350 & 8--1463 & 22,671,708 \\
    AI-READI$^{2}$ \cite{AIREADI-dataset,ai2024ai} & Controlled & 2024 & US & 1067  & 11 & 2--13 & 2,880,509 \\
    UCHTT1DM \cite{UCHTT1DM-dataset,langarica2024deep} & Open & 2024 & Chile & 20 & 6 & 3--8 & 29,174 \\
    CGMacros$^{3}$ \cite{CGMacros-dataset} & Open & 2025 & US & 45 & 11 & 8--20 & 629,825 \\
    \midrule
    Glucose-ML Total & & & & 2,559 & & & 38,008,810\\
    \bottomrule
  \end{tabular}
}
\vspace{1ex}
  \noindent\parbox{\textwidth}{\footnotesize $^{1}$OhioT1DM was initially released in year 2018, then it was updated to the current version released in year 2020.}
  \noindent\parbox{\textwidth}{\footnotesize $^{2}$This study leverages v2.0.0 of the AI-READI dataset comprising data from 1067 participants that was collected between July 19, 2023 and July 31, 2024.}
  \noindent\parbox{\textwidth}{\footnotesize $^{3}$The CGMacros dataset includes glucose values from both Dexcom and Freestyle Libre CGMs for each participant. This study leverages glucose values from only the Dexcom CGMs for analysis.}
\end{table}


The specific sensors used for collecting glucose data across all datasets include CGMs manufactured by Medtronic (i.e. Medtronic Enlite and Guardian) \cite{Medtronic-CGM}, Dexcom (i.e., Dexcom G5, G6, and Pro) \cite{Dexcom-CGM} and Abbott (i.e., FreeStyle Libre 1, 2 and Pro) \cite{FreeStyleLibre-CGM}. In general, CGM sensors by Medtronic and Dexcom repeatedly measure glucose every 5 minutes, and have a lifespan of up to 7 days and up to 10 days respectively, after which the sensor should be replaced. Meanwhile, Freestyle Libre CGMs by Abbott measure glucose every 15 minutes and have a lifespan of up to 14 days after which the sensor should be replaced. These CGMs are clinically-validated and FDA-approved sensors that report glucose readings between the range of 40 - 400 mg/dL \cite{danne2017international,battelino2019clinical}.

\subsection{Dataset accessibility}
Table \ref{tab:dataset_overview_update} shows that each individual dataset within the Glucose-ML collection is publicly available either through open or controlled access. The individual datasets are hosted on several hosting sites, including dedicated institutional sites \cite{OhioT1DM-dataset}, Vivli \cite{Vivli}, Physionet \cite{PhysioNet,goldberger2000physiobank}, Synapse \cite{Synapse}, Figshare \cite{FigShare}, Zenodo \cite{Zenodo}, FAIRhub \cite{FAIRhub}, and Github \cite{Github}.



\subsection{Ethics and fairness}
All datasets within the Glucose-ML collection are reported to be obtained in accordance with the relevant ethical guidelines, including approval by an appropriate institutional review board (IRB) and all participants provided informed consent. These datasets are fully de-identified such that they do not contain personally identifiable data. Additionally, a data use agreement (DUA) is required for datasets released via controlled access. Conversely, the majority of open access datasets are released with an appropriate license, such as the Open Data Commons Attribution License and the Creative Commons Attribution-NonCommercial License. The three exceptions are the UCHTT1DM \cite{UCHTT1DM-dataset} and ShanghaiT1DM \& ShanghaiT2DM \cite{ShanghaiDM-dataset} datasets which are released openly on Github and Figshare, but do not include an associated license. 


%% file: 4_comparative_analysis.tex
\section{Comparative analysis of diabetes datasets}
\begin{figure}
    \centering
    \includegraphics[width=\linewidth]{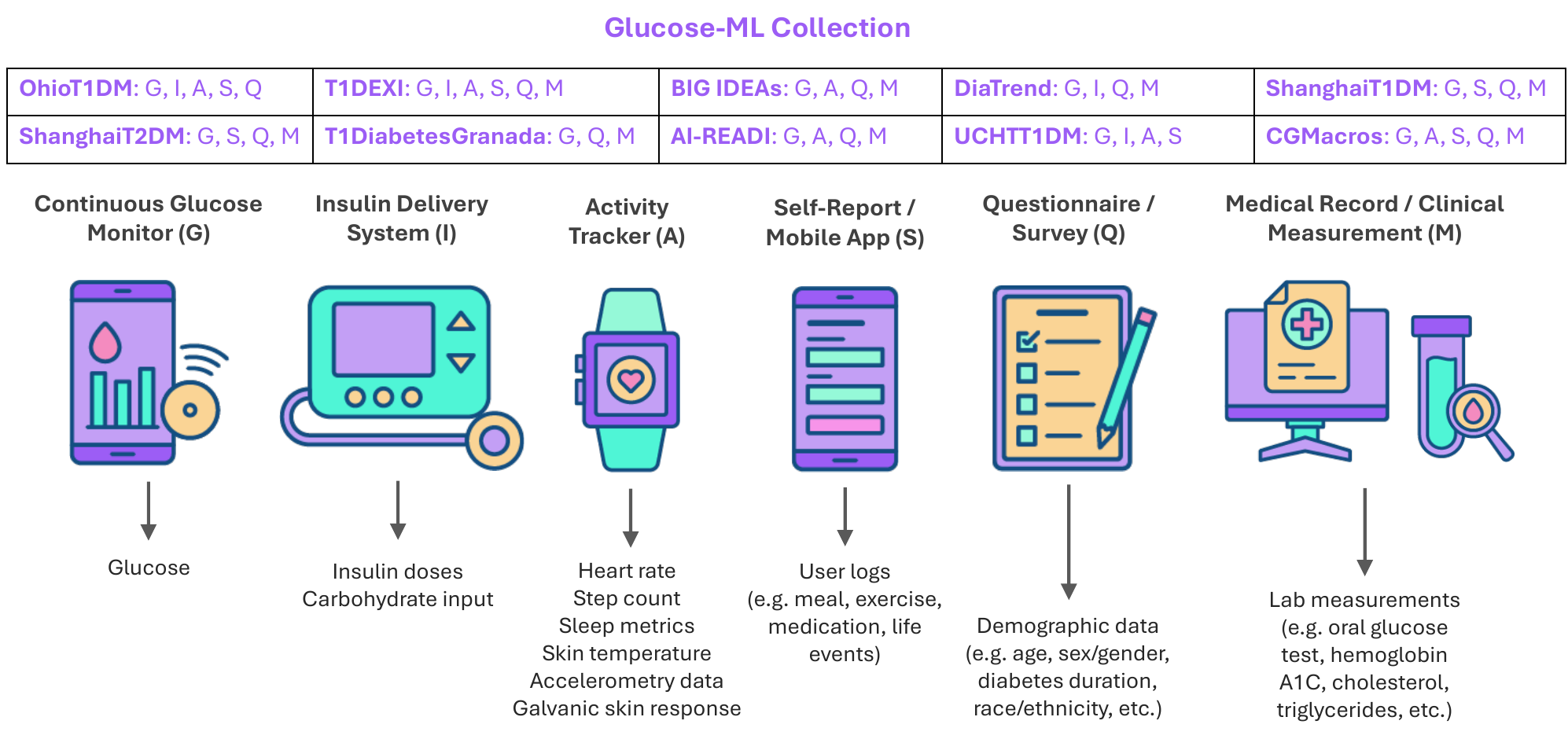}
    \caption{Overview of data types included within individual datasets in the Glucose-ML collection.}
    \label{fig:Glucose-ML-overview}
\end{figure}

In this study, we conducted comparative analysis across all 10 publicly-available datasets within the Glucose-ML collection to elicit unique strengths and weaknesses of each dataset. Our comparative analysis can also guide in the process of selecting diabetes-relevant datasets to support development of robust AI solutions. While the majority of our analysis focuses on glucose data from CGMs, it is important to note that these datasets also include diabetes-relevant data from other sources such as insulin delivery systems, activity trackers, user logs/mobile apps, questionnaires/surveys, and medical record/clinical measurements. Fig. \ref{fig:Glucose-ML-overview} presents a comparative overview of various data sources and data types across individual datasets in the Glucose-ML collection. From this figure, we can observe that the T1DEXI dataset is potentially the most comprehensive dataset as it includes glucose data, insulin-related data (e.g. insulin doses and carbohydrate input), activity tracker data (e.g. heart rate, step count, sleep metrics), user logs (e.g., meals, exercise), demographic data (e.g. age, sex/gender, diabetes duration), and lab measurements (e.g. hemoglobin A1C). However, T1DEXI comprise a population of people with T1D \textit{only}, and mean data duration is 27 days/person as shown in Table \ref{tab:dataset_overview_update}. 

\begin{figure}
    \centering
    \includegraphics[width=\linewidth]{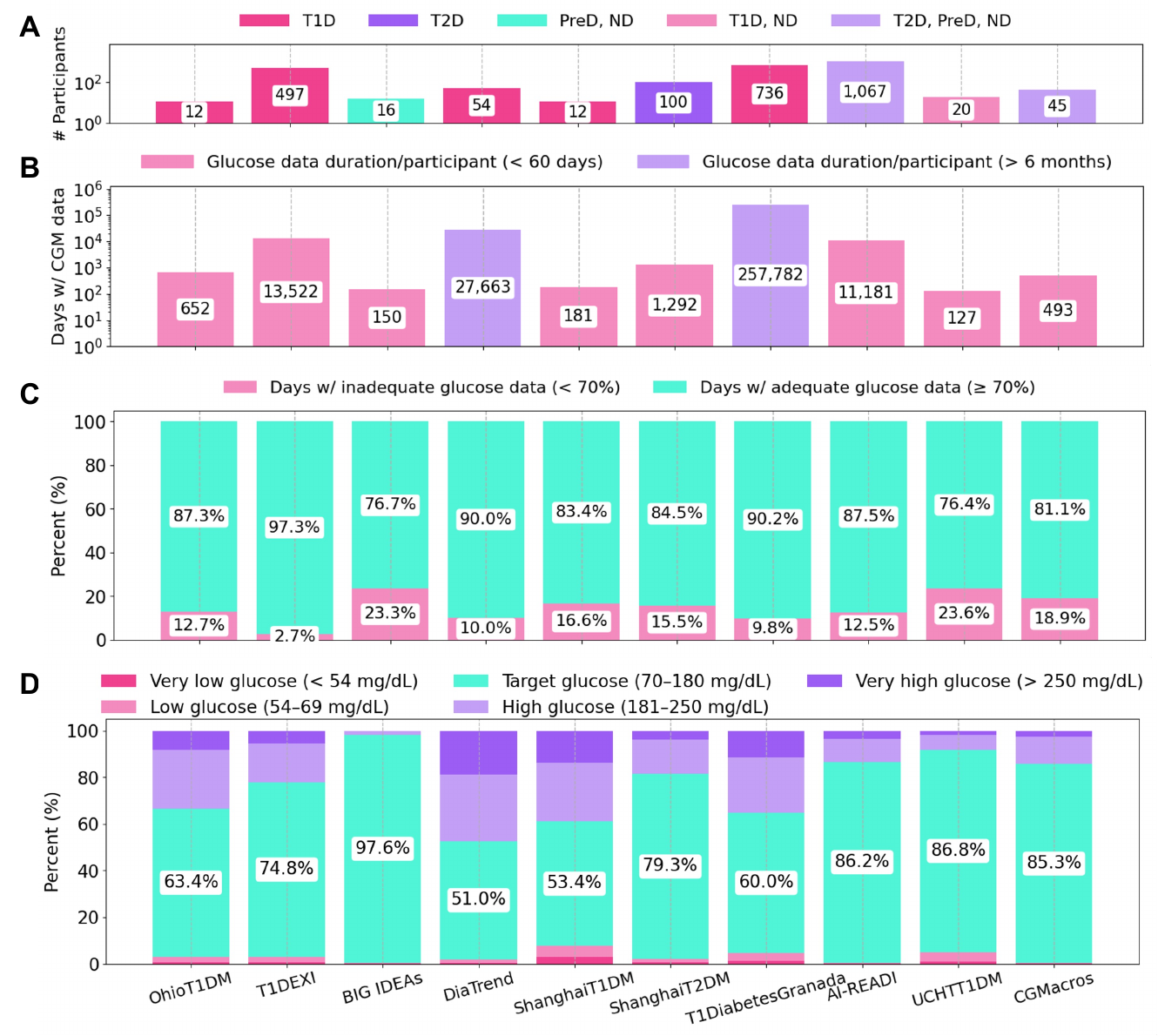}
    \caption{Comparative analysis of sample size and population (A), longitudinal glucose duration (B), data quality and sufficiency (C), and glucose dynamics (D) across 10 public diabetes datasets.}
    \label{fig:comparative-analysis}
\end{figure}

Fig. \ref{fig:comparative-analysis}A presents a comparison of the populations and sample size represented in each individual dataset within the Glucose-ML collection. From this figure, we can observe that there are between 12 - 1067 participants in each individual dataset. Five out of the ten datasets are solely from cohorts with T1D (50\%), one dataset is solely from a cohort with T2D (10\%), while the remaining four datasets include participants with T2D, prediabetes, and no diabetes (20\%), T1D and no diabetes (10\%) and prediabetes and no diabetes (10\%). From Fig. \ref{fig:comparative-analysis}B, we see that there are between 127 - 257,782 days of CGM data in each dataset. The majority of longitudinal diabetes datasets (80\%) include short durations of glucose data per participant (i.e., less than 60 days). Only DiaTrend \cite{prioleau2023diatrend} and T1DiabetesGranada \cite{T1DiabetesGranada-dataset} include long durations of glucose data per participant (i.e., more than 6 months, and in many cases, multiple years of longitudinal glucose data).

Given that glucose data in these longitudinal diabetes datasets is collected via wearable CGMs, missing data is not uncommon. Clinical consensus guidelines recommend the availability of more than 70\% of possible CGM readings to provide adequate glucose data for evaluating diabetes management \cite{battelino2019clinical,danne2017international}. Therefore, in this study we leverage insights from prior work to identify missing CGM data when glucose readings are not present for more than three times the expected sampling rate of the distinct CGM used within each dataset (i.e., Dexcom, Medtronic, or FreeStyle Libre CGMs) \cite{cui2025seasonal,fonda2013minding,vhaduri2020adherence}. For example, a Dexcom G6 CGM generally records glucose data every 5 minutes. Therefore, a period of missing data is identified when the difference between two consecutive timestamps of glucose readings has a duration of 15 minutes or more. Fig. \ref{fig:comparative-analysis}C presents an overview of the percentage of CGM days with adequate (i.e., $\geq$70\%) versus inadequate glucose data for each dataset. From this figure, we can observe that all datasets included in our collection have sufficient glucose data (i.e., < 25\% missing data) within the data collection period of each study. 

Finally, we compared the blood glucose dynamics of participants within each dataset using clinically-validated metrics, namely time in glucose ranges \cite{battelino2019clinical}. These metrics quantify the percentage of glucose readings that are within the target/normal range of 70 - 180 mg/dL (TIR), below the target range (TBR), and above the target range (TAR). Fig. \ref{fig:comparative-analysis}D presents an overview of the percentage of blood glucose values within the five clinically-relevant target ranges for each dataset. It is important to note that we only calculate the time in various glucose ranges on days with adequate glucose data (i.e., > 70\% of CGM data available). From Fig. \ref{fig:comparative-analysis}D, we can observe that DiaTrend \cite{prioleau2023diatrend} has the lowest TIR of 51\%, whereas the clinical recommendation for persons with T1D is to maintain a target TIR of greater than 70\% \cite{battelino2019clinical}. This significant difference between real-world blood glucose control in DiaTrend and the clinical recommendation shows that the DiaTrend dataset is the most dynamic and potentially the most challenging dataset to be used for developing and evaluating robust AI algorithms for populations with T1D. Conversely, the BIG IDEAs dataset \cite{bent2021engineering,BIGIDEAs-dataset} which was collected from a population with prediabetes and no diabetes has the highest TIR of 97.6\%. This observation shows that BIG IDEAs dataset comprises more stable and in-range glucose dynamics so AI algorithms developed and evaluated solely on this dataset will likely report better performance due to the less challenging context. 

%% file: 5_glucose_prediction.tex
\section{Case study: Blood glucose prediction}
Informed by several literature reviews within the field, we observed that blood glucose prediction is one of the most common task for which AI/ML algorithms are developed within the diabetes domain \cite{afsaneh2022recent,contreras2018artificial,felizardo2021data,ghimire2024deep,guan2023artificial,jacobs2023artificial,woldaregay2019data,xie2020benchmarking}. This is a classic time-series prediction problem for which statistical, traditional, and deep learning methods have been used to solve in prior literature. Given this, we conduct a case study for the task of predicting blood glucose 30-minutes ahead using two na\"{\i}ve baseline methods recommended to serve as benchmarks for comparison with new and more advanced AI algorithms. The objective of this case study is not to present a novel AI algorithm for the task, but to characterize and compare the performance of na\"{\i}ve prediction algorithms across 10 publicly available diabetes datasets within the GlucoseML collection. 

\subsection{Na\"{\i}ve Baseline Algorithms}
Building on best practice guidelines \cite{jacobs2023artificial}, we implement two na\"{\i}ve models, including 1) a zero-order hold predictor, and 2) a simple linear regression predictor, for the task of blood glucose prediction. 
\begin{itemize}
    \item \textbf{Zero-hold predictor} (also known as a persistent baseline) simply assumes the predicted value will not change in the future. Therefore, when using a zero-hold predictor for blood glucose prediction 30-mins ahead, we simply use the exact glucose value 30 minutes prior as the prediction 30 minutes ahead. This baseline implementation is described as linear extrapolation in the work by Fox et al. \cite{fox2018deep}. 
    \item \textbf{Simple linear regression predictor} fits a regression line across the most recent history of a given duration ($H$), then uses this regression line to predict blood glucose 30 minutes ahead \cite{jacobs2023artificial}. In our implementation of a simple linear regression predictor, we used a history of between 10 - 30 minutes based on the sampling frequency of the specific CGM used for data collection within each dataset. For studies that used a Dexcom or Medtronic CGM, both of which record approximately one glucose reading every 5 minutes, we used a history duration of 10 minutes with the goal of having three consecutive glucose readings at $t$, $t=5 mins$, and $t+10 mins$ as the basis for fitting a regression line. Conversely, for studies that used a FreeStyle Libre CGM (i.e., ShanghaiT1DM \& ShanghaiT2DM \cite{zhao2023chinese}, and T1DiabetesGranada \cite{rodriguez2023t1diabetesgranada}), we used a history duration of 30 minutes with the similar goal of having three consecutive glucose readings $t$, $t=15 mins$, $t+30 mins$ as the basis for fitting a regression line.
\end{itemize}

\subsection{Data Cleaning}
Toward the goal of predicting blood glucose 30 minutes ahead using the aforementioned na\"{\i}ve baseline algorithms, we started by conducting simple data cleaning steps to remove unrealistic glucose readings within individual datasets in the GlucoseML collection. Our first data cleaning step included removing non-numeric glucose readings and glucose readings that are not within the standard CGM range of 40 - 400mg/dL. Examples of excluded glucose recordings include recordings of "LOW" and "HIGH" in place of an actual glucose value in the AI-READI dataset. Following this, we removed consecutive glucose recordings with an unrealistic rate of change. Informed by the paper by DeSalvo and Buckingham, we removed consecutive glucose readings that were less than 30 seconds apart and/or that had a rate of change greater than 20mg/dL/min, which is 5 times the reported rate of change for detecting missed meal boluses \cite{desalvo2013continuous}. Table \ref{tab:total-glucose-samples-data-cleaning} in the Appendix provides a summary of the total glucose recordings in each dataset before the data cleaning steps, number of recordings excluded, and total glucose samples after data cleaning. 

\subsection{Algorithm Implementation, Handling Missing Data, and Performance Evaluation} Since our two na\"{\i}ve baseline algorithms have minimal or no parameters that need to be learned from a training dataset, our implementation of these methods was run on the full data duration of participants within the individual datasets (except for OhioT1DM). In the case of OhioT1DM \cite{marling2020ohiot1dm} where a train/test split was provided by the original data creators, we evaluated the na\"{\i}ve algorithms on the apportioned test set only. However, our implementation did not require splitting the other datasets into a training, validation, and/or test set.  Informed by best practice guidelines, we excluded periods of missing data from our analysis to avoid interpolation or extrapolation approaches that can introduce errors into the prediction model and reported accuracy \cite{jacobs2023artificial}. Finally, to facilitate easy comparison of the performance of na\"{\i}ve prediction algorithms across 10 publicly available diabetes datasets, we report the accuracy of each prediction method using root mean squared error (RMSE) \cite{hodson2022root,jacobs2023artificial}. A lower RMSE equates to less prediction error, while a higher RMSE equates to more prediction error and thus worse prediction performance. 

\textbf{Computing Resources}: All experiments were conducted on a MacBook Air (Apple M2 CPU, 8 GB and 16GB RAM), and run using CPUs only. The total estimated preprocessing time is $\approx$ 70 seconds. The runtime varied depending on the dataset duration, ranging from $\approx$ 1 minute to 4 hours per dataset. Per Table \ref{tab:dataset_overview_update} and Appendix Table \ref{tab:total-glucose-samples-data-cleaning}, we see that T1DiabetesGranada \cite{rodriguez2023t1diabetesgranada,T1DiabetesGranada-dataset} has the largest number of glucose samples (22.6 million) with an average of 1 year of glucose data per participant, and up to 4 years of glucose data across individual participants. As a result, this dataset required the longest runtime. The total estimated storage required for analysis in this study is $\approx$ 8 GB. 

\subsection{Results for Blood Glucose Prediction}
\begin{figure}
    \centering
    \includegraphics[width=\linewidth]{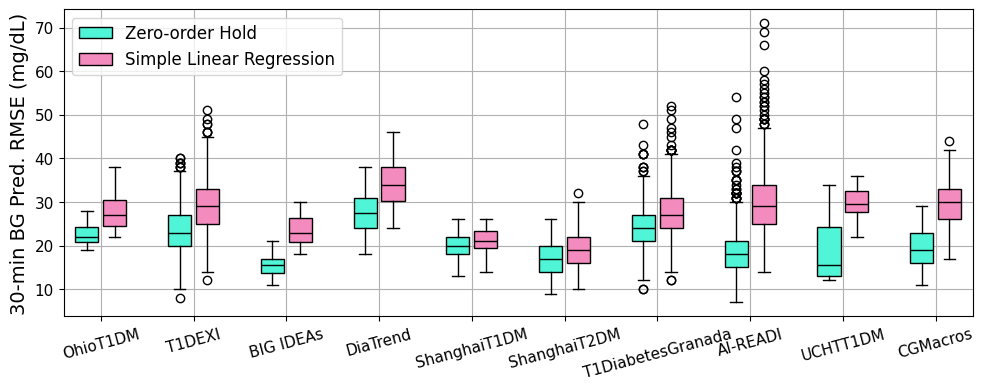}
    \caption{Performance overview for two na\"{\i}ve baseline algorithms, zero-order hold and simple linear regression, predicting blood glucose 30 minutes ahead using 10 publicly available diabetes datasets.}
    \label{fig:bg-pred-comparison}
\end{figure}

Fig. \ref{fig:bg-pred-comparison} and Table \ref{tab:glucose-prediction-results} present the results for predicting blood glucose 30 minutes ahead using the aforementioned na\"{\i}ve baseline methods. From Fig. \ref{fig:bg-pred-comparison}, we observe that the zero-order hold predictor performed better as evident by the lower RMSE compared to the simple linear regression predictor in all 10 diabetes datasets. Given this, we will focus the rest of our analysis on the results from the zero-order predictor for assessing the performance difference across datasets. From Table \ref{tab:glucose-prediction-results}, we observe that the zero-order hold predictor achieved the lowest RMSE of 16.1 $\pm$ 2.72 mg/dL on BIG IDEAs, meanwhile the same method achieved the highest RMSE of 28.14 $\pm$ 4.96 mg/dL on DiaTrend. A Mann-Whitney U test \cite{mcknight2010mann} was performed to compare the glucose prediction performance of the zero-order hold predictor on the BIG IDEAs and DiaTrend datasets. We found that there is a statistically significant difference in the glucose prediction performance of the same method on the BIG IDEAs dataset and the DiaTrend dataset; U-statistic = 8, Z= -5.92, $p$ = \num{3.16e-9}. These findings suggest that the dataset used for developing a blood glucose prediction algorithm significantly affects the performance obtained.

To ensure a more fair comparison of blood glucose prediction performance across datasets, we identified diabetes datasets with T2D, preD, and ND for comparison, and diabetes datasets with T1D only for comparison. We performed a Mann-Whitney U test to compare the glucose prediction performance on the BIG IDEAs and CGMacros datasets, both of which are more comparable in the represented population. Similarly, we found that there is a statistically significant difference in the glucose prediction performance on these two groups; U-statistic = 185, Z= -2.86, $p$ = 0.004. To also compare the prediction performance across T1D only datasets, we observed that the zero-order hold predictor achieved the lowest RMSE of 20.58 $\pm$ 3.39 mg/dL on the ShanghaiT1DM dataset and the highest RMSE of 28.14 $\pm$ 4.96 mg/dL on the DiaTrend dataset - see Table \ref{tab:glucose-prediction-results}. We also performed a Mann-Whitney U test to compare the glucose prediction performance on DiaTrend and ShanghaiT1DM datasets; both solely from T1D populations. We found that there is a statistically significant difference in the glucose prediction performance on these two groups; U-statistic = 584, Z= 4.33, $p$ = \num{1.6e-5}. These results further support that the dataset used for developing and evaluating a blood glucose prediction algorithm significantly affects the prediction performance obtained.

\begin{table}[]
\caption{Mean RMSE (std) for predicting blood glucose 30-minutes ahead using two naïve baselines. Lower RMSE means less prediction error while higher RMSE means more prediction error.}
\label{tab:glucose-prediction-results}
\centering
\begin{tabular}{lll}
\toprule
& \multicolumn{2}{c}{Mean RMSE (std) mg/dL}                   \\
    \cmidrule(r){2-3}
\begin{tabular}[c]{@{}l@{}}Publicly Available \\ Diabetes Datasets\end{tabular} & Zero-order Hold & Simple Linear Regression \\
\midrule

OhioT1DM \cite{marling2020ohiot1dm,OhioT1DM-dataset}                                    & 23.27 (2.92)             & 28.08 (4.80)                       \\
T1DEXI \cite{riddell2023examining,T1DEXI-dataset}                                      & 24.03 (5.38)             & 29.80 (6.28)                       \\
BIG IDEAs \cite{bent2021engineering,BIGIDEAs-dataset}                                   & 16.11 (2.72)             & 23.86 (3.91)                      \\
DiaTrend \cite{prioleau2023diatrend,DiaTrend-dataset}                                    & 28.14 (4.96)             & 35.00 (5.29)                       \\
ShanghaiT1DM \cite{zhao2023chinese,ShanghaiDM-dataset}                                & 20.58 (3.39)             & 21.70 (3.20)                        \\
ShanghaiT2DM \cite{zhao2023chinese,ShanghaiDM-dataset}                                & 17.51 (4.04)             & 19.82 (4.48)                      \\
T1DiabetesGranada \cite{rodriguez2023t1diabetesgranada,T1DiabetesGranada-dataset}                           & 24.81 (5.17)             & 28.08 (5.70)                       \\
AI-READI \cite{ai2024ai,AIREADI-dataset}                                    & 19.16 (4.96)             & 30.67 (7.06)                      \\
UCHTT1DM \cite{langarica2024deep,UCHTT1DM-dataset}                                    & 19.28 (7.10)              & 30.41 (4.00)                       \\
CGMacros \cite{CGMacros-dataset}                                    & 19.90 (4.68)              & 30.59 (5.64)     \\             
\bottomrule
\end{tabular}

\end{table}




%% file: 6_discussion.tex
\section{Discussion}

In this paper, we present Glucose-ML - a collection of 10 publicly available diabetes datasets to accelerate development and evaluation of transparent, reproducible, and robust AI solutions. This collection comprises longitudinal glucose and other data from 2500+ people living with diabetes (i.e., T1D and T2D) and people living without diabetes (i.e., prediabetes and no diabetes). The Glucose-ML collection includes over 300,000 days with CGM data and a total of 38 million glucose samples across all individual diabetes datasets. To our knowledge, the Glucose-ML collection is the largest and most comprehensive in literature to date. In addition to providing a review of recent publicly available diabetes dataset, this study includes comparative analysis of each dataset and a case study for blood glucose predicting using these datasets. Our results provide a benchmark for predicting blood glucose using each dataset within the Glucose-ML collection. Furthermore, our results show that the same AI algorithm can have significantly different results when developed and evaluated on different datasets.

Although this study did not include all public diabetes datasets available to facilitate AI development, we provide a more comprehensive review and analysis compared what has been presented in prior work. For example, Jacobs et al. \cite{jacobs2023artificial} present a review of current real-world diabetes datasets, but only include three datasets, namely OhioT1DM \cite{marling2020ohiot1dm}, T1DEXI \cite{riddell2023examining}, and Tidepool Big Data Donation \cite{TidepoolBigData}. The Tidepool Big Data Donation dataset has been used for data-centric analysis in a few studies \cite{bartolome2022computational,belsare2023understanding,mosquera2022incorporating,mosquera2019leveraging,prasanna2023hypoglycemia}, however, this dataset is not publicly available so it was not included in the Glucose-ML collection. Some examples of other open diabetes datasets include OpenAPS \cite{melmer2019glycaemic}, REPLACE-BG \cite{aleppo2017replace} and Maastricht study \cite{schram2014maastricht,van2021machine}. The OpenAPS dataset has been used for data-centric analysis and developing AI solutions in prior work \cite{hameed2020comparing,shahid2022large}. However, this dataset is collected from a highly selective, technology-adept population of individuals with T1D who use a fully-closed loop automated insulin delivery (AID) system for diabetes management so it was not included in the Glucose-ML collection \cite{melmer2019glycaemic}. Meanwhile, the REPLACE-BG \cite{aleppo2017replace} (collected in year 2015 - 2016) and Maastricht study \cite{schram2014maastricht,van2021machine} (initially reported in 2014) datasets were not included in the Glucose-ML collection because they are not recent and were collected/released before year 2018.








\subsection{Recommendations for developing robust AI solutions for diabetes and beyond}


Comprehensive recommendation on developing robust AI algorithms is beyond the scope of this work, so we refer readers to related work \cite{gundersen2023,jacobs2023artificial,lekadir2025future}. Here, we present three recommendations that build on findings from this study.

\begin{itemize}
    \item \textbf{Data Selection}: A best practice guideline for AI development and evaluation is to leverage multiple datasets with representative subgroups and representative variability for the target population and task. For diabetes-centered AI solutions, the selected datasets should include individuals with representative variation in glycemic control, hemoglobin A1C, age groups, geographical locations, racial/ethnic groups, gender, and more - see Table \ref{tab:dataset_overview_update}, \ref{tab:detailed-overview-diabetes-datasets} and \ref{tab:BGcontrol-metrics}.
    \item \textbf{Model Design}: Newly proposed AI models should \textit{always} be compared with a na\"{\i}ve baseline model. Several studies in literature do not include a baseline (e.g. \cite{lee2023glucose,rajagopal2023novel,carvalho2024glucose,xiong2025exploring}) or only include regression-based baselines or other more advanced baselines (e.g. \cite{hameed2020comparing,gu2020neural,zhu2022iomt,zhu2022personalized}). A best practice guideline is to include a zero-order hold predictor as a na\"{\i}ve baseline for blood glucose prediction tasks in addition to any other baselines. This is especially critical when a private dataset is used for AI development and evaluation. 
    \item \textbf{Model Evaluation}: To enable comparison with related work, it is critical for newly proposed AI solutions to \textit{not only} be developed with simulated and/or private datasets (e.g. \cite{langarica2023meta,mohebbi2020short,kim2022intelligent,li2019convolutional}). A best practice guideline is to develop and evaluate AI solutions on publicly available datasets, in addition to any private datasets. Additionally, models should not be evaluated on test sets that have been smoothed or that include interpolation of missing data \cite{jacobs2023artificial}. These common pitfalls can lead to invalid estimates of accuracy.
    
\end{itemize}

\subsection{Ethical considerations}

\textbf{Representation across the Glucose-ML collection}: We assess representation of the Glucose-ML collection across three dimensions: sex/gender, age, and race/ethnicity. With regard to sex/gender, Table \ref{tab:detailed-overview-diabetes-datasets} in the Appendix shows that the Glucose-ML collection includes a fair representation across sex/gender (total: 1077 males (42\%) and 1482 females (58\%)). With regards to age,  the Glucose-ML collection does not include representation of children and/or adolescents - see Table \ref{tab:detailed-overview-diabetes-datasets}. However, T1D and increasing incidences of prediabetes and T2D are not uncommon in pediatric populations \cite{lawrence2021trends,nobili2019prevalence,zhang2023global}. With regard to race/ethnicity, research shows that diabetes is more prevalent across certain racial/ethnic groups \cite{cheng2019prevalence}. For example, in US adults, diabetes is more prevalent amongst American Indian and Alaska Native adults, followed by non-Hispanic Black adults, and adults of Hispanic origin \cite{CDC-diabetes-statistics}. However, data on race/ethnicity is only reported in a few datasets (i.e., T1DEXI \cite{T1DEXI-dataset}, DiaTrend \cite{DiaTrend-dataset}, AI-READI \cite{AIREADI-dataset}, and CGMacros \cite{CGMacros-dataset}. Of the four datasets (40\%) that report race/ethnicity, T1DEXI \cite{riddell2023examining,T1DEXI-dataset} and DiaTrend \cite{prioleau2023diatrend,DiaTrend-dataset} have a large representation of non-Hispanic Whites/Caucasians; 87\% and 89\%, respectively. CGMacros has a large representation of Hispanics/Latinos (75.5\%) \cite{CGMacros-dataset}. While AI-READI has a more balanced representation of persons from Hispanic (21\%), Asian (23\%), Black (27\%), and White (29\%) race/ethnic groups \cite{AIREADI-dataset}. 


\textbf{Intended Data Use:} The Glucose-ML collection has been curated to accelerate research and development of AI solutions for diabetes and beyond. It is imperative to respect the privacy of participants who have openly shared their data with qualified researchers/innovators. As a result, data users should not attempt to re-identify participants, including for re-identification theory research.






\subsection{Limitations}
Albeit the strengths of this work, there are some limitations that should be acknowledged. Firstly, this paper focuses solely on evaluating longitudinal glucose data from publicly available diabetes datasets. Hence, we did not delve into other longitudinal/diabetes-related data streams within each dataset, such as insulin-related data (e.g. insulin and carbohydrate input), activity tracker data (e.g. heart rate, step count, and sleep metrics), user logs (e.g. medication, meals, and life events), and clinical measurements (e.g. hemoglobin A1C and cholesterol). Secondly, despite our faithful attempt to provide a review and comparative analysis of recent open diabetes datasets, there is a chance that we unintentionally omitted some relevant datasets. For example, two recent T1D datasets, D1NAMO \cite{dubosson2018open,D1NAMO-dataset} and T1DEXIP \cite{T1DEXIP-dataset,riddell2024acute}, were not included in the Glucose-ML collection. When writing this manuscript, the authors were awaiting access to T1DEXIP and had not discovered D1NAMO.

%% file: 9_appendix.tex
\newpage
\section{Technical Appendices and Supplementary Material}

\begin{table}[h]
  \caption{Overview of the population, demographics and hemoglobin A1C for diabetes datasets within the Glucose-ML collection.}
  \label{tab:detailed-overview-diabetes-datasets}
  \centering
  \resizebox{\textwidth}{!}{%
  \begin{tabular}{llllllll}
    \toprule
     \multirow{2}{*}{\shortstack{Publicly available \\Diabetes Datasets}} & & & Gender & \multicolumn{2}{c}{Age (years)} & \multicolumn{2}{c}{Hemoglobin A1C (\%)}$^{2}$ \\
    \cmidrule(r){5-6} \cmidrule(r){7-8}
     & Population & Participants & M / F & Mean (SD) & Range & Mean (SD) & Range \\
    \midrule
    OhioT1DM & T1D & 12 &  7 / 5 & NR & 20--80 & NR & NR \\
    T1DEXI & T1D & 497 & 134 / 363 & 36.7 (14.0) & 18--70 & 6.6 (0.8) & 4.8--10.0 \\
    BIG IDEAs & PreD \& ND & 16 & 7 / 9 & NR & 35--65 & 5.7 (0.3) & 5.3--6.4 \\
    DiaTrend & T1D & 54 & 17 / 37 & 32.1 (15.0)$^{1}$ & 19--70 & 7.8 (0.8) & 6.3--10.0 \\
    ShanghaiT1DM & T1D & 16 & 5 / 7 & 57.8 (11.1) & 37--73 & 8.2 (3.4) & 4.9--15.1 \\
    ShanghaiT2DM & T2D & 100 & 56 / 44 & 60.2 (13.7) & 22--97 & 6.9 (2.5) & 2.1--13.3 \\
    T1DiabetesGranada & T1D & 736 & 363 / 373 & 43.6 (16.1) & 14--85 & 8.3 (1.8) & 4.0--18.0 \\
    AI-READI & T2D \& PreD / ND & 703 / 364 & 462 / 605 & 60.3 (11.1) & 40--87 & NR & NR \\
    UCHTT1DM & T1D / ND & 9 / 11 & 10 / 10 & 27.2 (4.0) & NR & NR & 4.4--8.2 \\
    CGMacros & T2D / PreD / ND & 14 / 16 / 15 & 16 / 29 & 48.1 (12.7) & 18--69 & 6.1 (0.9) & 4.6--8.5 \\
    \bottomrule
  \end{tabular}
  }
  \vspace{1ex}
  \noindent\parbox{\textwidth}{\footnotesize $^{1}$ Seventeen participants in DiaTrend reported their age within a range; midpoint values in the age range were used to calculate summary statistics. }
  \noindent\parbox{\textwidth}{\footnotesize $^{2}$ For participants with multiple HbA1c records in ShanghaiT1DM, ShanghaiT2DM, and T1DiabetesGranada, only the first HbA1C measurement was used to obtain summary statistics. }
\end{table}

\begin{table}[h]
  \caption{Overview of glycemic control across diabetes datasets within the Glucose-ML collection.}
  \label{tab:BGcontrol-metrics}
  \centering
  \resizebox{\textwidth}{!}{%
  \begin{tabular}{lllllll}
    \toprule
    \multirow{2}{*}{\shortstack{Publicly available \\Diabetes Datasets}} &
    \multicolumn{2}{c}{TIR per Day (\%)} &
    \multicolumn{2}{c}{Avg Glucose per Day (mg/dL)} &
    \multicolumn{2}{c}{Glycemic Variability per Day (\%)} \\
    \cmidrule(lr){2-3} \cmidrule(lr){4-5} \cmidrule(lr){6-7}
     & Mean (SD) & Range & Mean (SD) & Range & Mean (SD) & Range \\
    \midrule
    OhioT1DM & 63.2 (20.7) & 0--100 & 160.1 (31.1) & 99.4--277.6 & 31.5 (8.5) & 11.7--57.2 \\
    T1DEXI & 74.7 (20.0) & 0--100 & 146.1 (32.1) & 66.4--365.9 & 29.2 (8.4) & 6.6--80.0 \\
    BIG IDEAs & 97.7 (3.8) & 79.1--100 & 114.1 (11.6) & 86.2--135.5 & 16.5 (5.2) & 5.3--37.8 \\
    DiaTrend & 50.7 (22.4) & 0--100 & 186.5 (41.2) & 70.3--397.2 & 32.0 (8.7) & 3.2--91.3 \\
    ShanghaiT1DM & 53.4 (22.6) & 0--100 & 165.0 (42.9) & 57.2--257.1 & 33.3 (10.4) & 12.9--78.3 \\
    ShanghaiT2DM & 79.2 (21.9) & 0--100 & 139.9 (33.9) & 40.6--285.9 & 24.7 (7.8) & 6.4--55.8 \\
    T1DiabetesGranada & 59.9 (23.6) & 0--100 & 164.1 (42.1) & 48.9--493.7 & 32.7 (10.0) & 3.3--101.1 \\
    AI-READI & 85.7 (23.6) & 0--100 & 139.0 (39.7) & 47.5--395.3 & 17.7 (6.3) & 1.3--60.1 \\
    UCHTT1DM & 86.5 (16.7) & 26.8--100 & 110.6 (30.8) & 71.3--200.8 & 21.2 (11.6) & 6.6--62.6 \\
    CGMacros & 85.3 (20.9) & 0--100 & 141.1 (29.4) & 88.6--257.1 & 18.7 (6.4) & 7.1--42.3 \\
    \bottomrule
  \end{tabular}
  }
\end{table}

\begin{table}[h]
  \caption{Total glucose samples before and after data cleaning.}
  \label{tab:total-glucose-samples-data-cleaning}
  \centering
  \resizebox{\textwidth}{!}{%
  \begin{tabular}{llll}
    \toprule
     \multirow{2}{*}{\shortstack{Publicly available \\Diabetes Datasets}} & \multicolumn{3}{c}{Total Glucose Samples} \\
    \cmidrule(lr){2-4}
      & Before Data Cleaning & Excluded w/ Data Cleaning & After Data Cleaning \\
    \midrule
    OhioT1DM & 166,533 & 3 & 166,530 \\
    T1DEXI & 3,785,253 & 6,728 & 3,778,525 \\
    BIG IDEAs & 36,898 & 0 & 36,898 \\
    DiaTrend & 7,680,740 & 115,026 & 7,565,714 \\
    ShanghaiT1DM & 15,695 & 0 & 15,695 \\
    ShanghaiT2DM & 112,475 & 13 & 112,462 \\
    T1DiabetesGranada & 22,671,708 & 62,956 & 22,608,752 \\
    AI-READI & 2,880,509 & 20,159 & 2,860,350 \\
    UCHTT1DM & 29,174 & 0 & 29,174 \\
    CGMacros & 629,825 & 0 & 629,825 \\
    \bottomrule
  \end{tabular}
  }
\end{table}

\begin{table}[]
\caption{The 30-minutes blood glucose prediction RMSE using two na\"{\i}ve baselines. This table shows the mean RMSE (std) for predicting all glucose values within each dataset (i.e., overall) and when the ground-truth glucose values is within 3 clinically-relevant ranges: 1) TBR (BG < 70 mg/dL), 2) TIR (70 < BG < 180mg/dL), and 3) TAR (BG > 180 mg/dL). Lower RMSE means less prediction error.}
\begin{tabular}{llll}
\toprule
& & \multicolumn{2}{c}{Mean RMSE (std) mg/dL}                   \\
    \cmidrule(r){3-4}
\begin{tabular}[c]{@{}l@{}}Publicly Available\\ Diabetes Datasets\end{tabular} &                & Zero-order Hold & Simple Linear Regression \\
\midrule
OhioT1DM                                                                                & Overall        & 23.27 (2.92)             & 28.08 (4.80)                       \\
                                                                                        & $<$ 70 mg/dL   & 16.95 (3.67)             & 19.88 (6.92)                      \\
                                                                                        & 70 - 180 mg/dL & 21.14 (2.75)             & 25.38 (4.61)                      \\
                                                                                        & $>$ 180 mg/dL  & 27.58 (4.89)             & 32.69 (8.31)                      \\ \hline
T1DEXI                                                                                  & Overall        & 24.03 (5.38)             & 29.80 (6.28)                       \\
                                                                                        & $<$ 70 mg/dL   & 29.04 (9.79)             & 30.99 (11.29)                     \\
                                                                                        & 70 - 180 mg/dL & 22.17 (4.71)             & 28.37 (5.80)                       \\
                                                                                        & $>$ 180 mg/dL  & 30.60 (6.73)              & 35.05 (7.81)                      \\ \hline
BIG IDEAs                                                                               & Overall        & 16.11 (2.72)             & 23.86 (3.91)                      \\
                                                                                        & $<$ 70 mg/dL   & 27.30 (14.24)             & 45.07 (29.46)                     \\
                                                                                        & 70 - 180 mg/dL & 15.20 (2.16)              & 23.17 (3.86)                      \\
                                                                                        & $>$ 180 mg/dL  & 43.74 (7.33)             & 43.53 (15.51)                     \\ \hline
DiaTrend                                                                                & Overall        & 28.14 (4.96)             & 35.00 (5.29)                       \\
                                                                                        & $<$ 70 mg/dL   & 30.73 (8.81)             & 33.45 (9.49)                      \\
                                                                                        & 70 - 180 mg/dL & 24.99 (4.38)             & 31.76 (5.13)                      \\
                                                                                        & $>$ 180 mg/dL  & 31.82 (5.27)             & 38.86 (5.49)                      \\ \hline
ShanghaiT1DM                                                                            & Overall        & 20.58 (3.39)             & 21.70 (3.20)                        \\
                                                                                        & $<$ 70 mg/dL   & 14.78 (10.32)            & 14.65 (5.96)                      \\
                                                                                        & 70 - 180 mg/dL & 19.06 (4.90)              & 19.79 (4.34)                      \\
                                                                                        & $>$ 180 mg/dL  & 24.10 (4.05)              & 24.92 (4.44)                      \\ \hline
ShanghaiT2DM                                                                            & Overall        & 17.51 (4.04)             & 19.82 (4.48)                      \\
                                                                                        & $<$ 70 mg/dL   & 13.85 (7.11)             & 15.34 (8.21)                      \\
                                                                                        & 70 - 180 mg/dL & 15.05 (3.46)             & 17.44 (4.43)                      \\
                                                                                        & $>$ 180 mg/dL  & 28.39 (9.20)              & 31.00 (9.97)                       \\ \hline
T1DiabetesGranada                                                                       & Overall        & 24.81 (5.17)             & 28.08 (5.70)                       \\
                                                                                        & $<$ 70 mg/dL   & 19.30 (7.32)              & 22.24 (8.39)                      \\
                                                                                        & 70 - 180 mg/dL & 23.14 (5.15)             & 26.82 (6.04)                      \\
                                                                                        & $>$ 180 mg/dL  & 28.81 (5.55)             & 31.45 (5.99)                      \\ \hline
AI-READI                                                                                & Overall        & 19.16 (4.96)             & 30.67 (7.06)                      \\
                                                                                        & $<$ 70 mg/dL   & 33.86 (19.62)            & 42.96 (25.08)                     \\
                                                                                        & 70 - 180 mg/dL & 17.34 (6.45)             & 28.85 (8.16)                      \\
                                                                                        & $>$ 180 mg/dL  & 40.41 (15.76)            & 47.80 (17.17)                      \\ \hline
UCHTT1DM                                                                                & Overall        & 19.28 (7.10)              & 30.41 (4.00)                       \\
                                                                                        & $<$ 70 mg/dL   & 21.90 (6.35)              & 28.33 (8.37)                      \\
                                                                                        & 70 - 180 mg/dL & 18.20 (6.27)              & 29.63 (4.03)                      \\
                                                                                        & $>$ 180 mg/dL  & 31.56 (9.21)             & 37.80 (6.71)                       \\ \hline
CGMacros\_Dexcom                                                                        & Overall        & 19.90 (4.68)              & 30.59 (5.64)                      \\
                                                                                        & $<$ 70 mg/dL   & 33.75 (18.90)             & 45.66 (24.76)                     \\
                                                                                        & 70 - 180 mg/dL & 16.88 (2.60)              & 27.68 (5.13)                      \\
                                                                                        & $>$ 180 mg/dL  & 41.75 (11.90)             & 48.81 (12.87)                     \\
\bottomrule
\end{tabular}
\end{table}